\documentclass{article} %
\usepackage{spconf,times}

\usepackage{amsmath,amsfonts,bm}

\def\eqref#1{equation~\ref{#1}}

\def\1{\bm{1}}

\def\vtheta{{\bm{\theta}}}

\def\vx{{\bm{x}}}
\def\vy{{\bm{y}}}
\def\vz{{\bm{z}}}

\DeclareMathAlphabet{\mathsfit}{\encodingdefault}{\sfdefault}{m}{sl}
\SetMathAlphabet{\mathsfit}{bold}{\encodingdefault}{\sfdefault}{bx}{n}

\newcommand{\norm}[1]{\left\lVert#1\right\rVert}
\def\vzh{{\bm{\hat{z}}}}
\def\vzt{{\bm{\tilde{z}}}}

\usepackage{hyperref}
\usepackage{url}
\usepackage{multirow}
\usepackage{booktabs}
\usepackage{tikz}
\usetikzlibrary{shapes,snakes}
\usepackage{xcolor}
\usepackage[ruled,vlined]{algorithm2e}

\usepackage[colorinlistoftodos, textwidth=20mm]{todonotes}
\newif\ifdraftydraft
\draftydrafttrue

\title{Joint Masked CPC and CTC Training for ASR}

\name{Chaitanya Talnikar, Tatiana Likhomanenko, Ronan Collobert, Gabriel Synnaeve}
\address{Facebook AI Research, New York, Menlo Park \& Paris, USA \& France}

\begin{document}

\maketitle

\begin{abstract}
Self-supervised learning (SSL) has shown promise in learning representations of audio that are useful for automatic speech recognition (ASR). But, training SSL models like wav2vec~2.0 requires a two-stage pipeline. In this paper we demonstrate a single-stage training of ASR models that can utilize both unlabeled and labeled data. During training, we alternately minimize two losses: an unsupervised masked Contrastive Predictive Coding (CPC) loss 
and the supervised audio-to-text alignment loss Connectionist Temporal Classification (CTC). We show that this joint training method directly optimizes performance for the downstream ASR task using unsupervised data while achieving similar word error rates to wav2vec~2.0 on the Librispeech 100-hours dataset. 
Finally, we postulate that solving the contrastive task is a regularization for the supervised CTC loss. 
\end{abstract}

\section{Introduction}
Deep learning has been impactful in building state-of-the-art end-to-end speech recognition systems \cite{amodei2016deep,synnaeve2020endtoend,park2020improved}. But, they 
typically require large amounts of annotated speech data in the form of transcripts. Whereas, humans
are able to learn language and speech with little supervision. 

Recently, self-supervised learning (SSL) has been proposed as a method for training automatic speech recognition (ASR) models
by pre-training on large amount of unlabeled data and then fine-tuning the speech recognition model on labeled data,
for example contrastive
predictive coding (CPC)~\cite{oord2018representation}. 
While these methods \cite{kharitonov2020data,librilight}
have achieved impressive results on low-resource speech datasets, their goal
is to learn speech representations that are useful for multiple speech-related tasks. 
Training an ASR model using SSL methods is a two-stage process as it
requires running separate pre-training and fine-tuning experiments and 
jointly tuning hyperparameters for both stages.
It is unclear how much pre-training is required to achieve reasonable performance on the downstream task of speech recognition.

In this paper, we propose a training method for ASR models that combines SSL and supervised learning
in a single stage. The model is trained by jointly minimizing a loss on labeled data, and a loss on unlabeled data. The supervised loss is the Connectionist Temporal Classification (CTC) loss~\cite{graves2006connectionist}, while the unsupervised loss is based on  a masked variant of CPC. As both losses are optimized jointly, our method allows early stopping by measuring the performance of the model for the downstream task on the validation dataset.

We show that a model trained using our method (with no quantization) achieves equivalent word error rate (WER) when trained on 960-hours of unlabeled data and 100-hours of labeled data to a model that is trained using the two-stage process of wav2vec~2.0~\cite{baevski2020wav2vec} (with quantization), which is a method based on masked CPC. Additionally,
we verify that our method provides a regularization to the supervised loss when only using labeled
data.

\section{Joint Training}
We propose to train our speech recognition model in a single stage, by jointly minimizing a supervised and an unsupervised loss.
Our training procedure alternates between minimizing the unsupervised loss on unlabeled data and minimizing the supervised loss on labeled data.

\begin{algorithm}[t!]
\SetAlgoLined
\KwData{Labeled data $L = \{\vx, \vy\}$, Unlabeled data $U = \{\vx\}$ }
\KwResult{Acoustic model $p_\vtheta$}
Randomly initialize parameters of the acoustic model~$p_\vtheta$\;
  \Repeat{convergence in word error rate or maximum iterations are reached}{
    \Repeat{$N$ times for $\vx\in U$}{
        1. Forward the model with Eq. (\ref{eq:encoder}) and (\ref{eq:context}) obtaining $\vz$ and $\vzt$ \\
        2. Compute $g_u = \nabla_\vtheta \mathcal{L}_u(\vtheta, \vx)$ using $\vz,\,\vzt$ \\
        3. Update $p_\vtheta$ with $\eta_u$ and $g_u$ \\
    }
    4. Forward the model for $\vx\in L$ with Eq.~(\ref{eq:encoder})-(\ref{eq:linear}) obtaining $p_\vtheta(\vy|\vx)$ \\
    5. Compute $g_s = \nabla_\vtheta \mathcal{L}_s(\vtheta, \vx, \vy)$ using $p_\vtheta(\vy|\vx)$\\
    6. Update $p_\vtheta$ with $\eta_s$ and $g_s$ \\
  }
 \caption{Alternating minimization algorithm.}
\label{algorithm}
\end{algorithm}

\subsection{Model}

Our model is a neural network architecture which gets as input raw audio ($\vx$) and outputs token ($\vy$) probabilities $p_\vtheta(\vy_t|\vx)$ at time $t$ with the following functions:
\begin{align}
    \label{eq:encoder}
    \vz &= f(\vx) \\
    \label{eq:context}
    \tilde{\vz} &= g(\text{mask}(\vz))\\
    p_\vtheta(\vy|\vx) &= h(\tilde{\vz}).
    \label{eq:linear}
\end{align}
where a convolutional encoder $f: X \to Z$ maps raw input audio into features at $20$ms stride with a receptive field $30$ms. These encoder features $\vz$ (with optional masking of certain frames) are passed as input into a transformer-based~\cite{vaswani2017attention} context network with full attention $g: Z \to \tilde{Z}$. 
Finally, the context features $\tilde{\vz}$ are used to generate output token probabilities $p_\vtheta(\vy_t|\vx)$ at time frame $t$ using a linear layer and softmax non-linearity $h: \tilde{Z} \to Y$.

\subsection{Unsupervised and supervised losses}
The supervised loss is CTC~\cite{graves2006connectionist}, denoted as $\mathcal{L}_s(\vtheta, \vx, \vy)$ in the paper.
The unsupervised loss is the self-supervision loss used for pre-training in wav2vec \cite{baevski2020wav2vec}. This loss can be viewed as a contrastive predictive coding \cite{oord2018representation} loss, where the task is to predict the masked encoder features~\cite{devlin2018bert} rather than predicting future encoder features given past encoder features.
In this loss, a certain percentage of the encoder features $\vz$ (controlled by the masking probability) are masked 
at time frames $t_{i_1}, t_{i_2}, ... , t_{i_T}$, where $i_1, i_2, ... , i_T$ denote
the masking indices. The features, for example $\vz_{t_{i_1}}$, are masked by replacing it with a learnt mask embedding.
The masked encoder features $\vzh=\text{mask}(\vz)$ are passed as input to the context network, which is responsible for reproducing the features $\vz$. 
The accuracy of reproduction is measured using a contrastive loss by comparing the similarity between the predicted features $\tilde{\vz}$ from the context network at masked indices (anchor) and the input features $\vz$ of the context network at masked indices (positive sample) against other encoder features at non-masked indices (negative samples): 
\begin{equation}
    \mathcal{L}_u(\vtheta, \vx) = \frac{1}{T}\sum_t -\mathrm{log}
          \frac{s\big(\vz_t, \vzt_t\big)}
          {s\big(\vz_t, \vzt_t\big) + 
          \sum_{t^\prime}s\big(\vz_{t^\prime}, \vzt_t\big)}
\end{equation}
where $s(\vz_t, \vzt_t) = \frac{1}{\tau}\mathrm{exp}(\frac{\vz_t\cdot\vzt_t}{\norm{\vz_t}\norm{\vzt_t}})$. The time frame $t$ denotes the index of the $T$ masked features, $\vz_{t^\prime}$ are encoder features sampled from time frames $t^\prime$ other than time frame $t$, $\tau$ is a tunable hyperparameter called temperature.

\subsection{Alternate minimization}
The model is trained by alternately minimizing the two losses. Using a minibatch from the unlabeled data, the gradient of the unsupervised loss is used to update the model parameters for $N$ steps, followed by the gradient of the supervised loss (using a minibatch from labeled data) for $1$ step. 
This process is repeated until convergence of the word error rate on the validation dataset. A brief description is shown in Algorithm~\ref{algorithm}.

Separate adaptive momentum optimizers are used for each of the two losses with different learning rates: $\eta_u$ for the unsupervised loss and $\eta_s$ for the supervised loss. The
two optimizers maintain their state independently, while sharing the parameters of the model. 
This ensures that the momentum averaging for one loss
is not affected by the gradient updates from the other loss, leading to faster convergence. 
Experiments with a single optimizer show worse performance on the downstream task compared
to the usage of two optimizers.

The ratio of unsupervised to supervised loss updates, $N$:1, is chosen to be 1:1. This
results in equal opportunity for the unsupervised and supervised tasks to affect the weights
of the network as a function of the total number of updates. Choosing an update ratio that
favors the unsupervised task results in a more computationally expensive training. While,
an update ratio that is biased towards the supervised task produces an ASR model that does not improve
over a supervised baseline. 

The learning rate ratio is biased towards the unsupervised task as compared to the supervised task.
Using a learning rate ratio of 1:1 or one that favors the supervised task results in an ASR model that does not
improve over a supervised baseline.
\section{Experimental Setup}
\subsection{Datasets}
The experiments use the Librispeech~\cite{panayotov2015librispeech} 960-hours dataset 
as the unsupervised dataset. The supervised dataset is a subset of Librispeech: either 100-hours or 960-hours (full). During training, samples in the dataset that are smaller than 2 seconds or longer than 33 seconds are filtered out. The performance of the trained model is validated on the dev-clean/other datasets of Librispeech and tested on the test-clean/other datasets.

\subsection{Architecture details}
Similar to wav2vec 2.0 \cite{baevski2020wav2vec}, the convolutional encoder network consists of a stack of 7 convolutions with kernel size $(10,3,3,3,3,2,2)$ and strides $(5,2,2,2,2,2,2)$ respectively. The number of input and output channels in the convolution is $512$. Additionally, the input audio is normalized in the time dimension before it is passed into the convolutional encoder. 

We use two versions of the model, \textsc{Base} and \textsc{Large}.
The transformer context network for the \textsc{Base} model is composed of a convolutional relative positional embedding layer with kernel size 128 and group size 16, followed by a stack of 12 transformer layers with 8 heads. The hidden dimension is 768 and the feed-forward network dimension is 3072. Each transformer layer uses layer dropout~\cite{fan2019reducing} with probability 0.05 and dropout with probability 0.1.
The transformer context network for the \textsc{Large} model uses a stack of 24 transformer layers with 16 heads. The hidden dimension is 1024 and the feed-forward network dimension is 4096. Each transformer layer uses layer dropout with probability 0.2 and dropout with probability 0.1.
The linear classifier is trained to output letter-based tokens, which consist of 26 English alphabet letters, augmented with the apostrophe and a word boundary token. The total number of parameters for the \textsc{Base} model is 94.3M and the \textsc{Large} model is 315M. 
The masking probability is $0.075$ for the \textsc{Base} model and $0.065$ for the \textsc{Large} model. The number of masked tokens per sample is $10$. The number of negative samples used in the contrastive loss is 100 and the temperature is 0.1. A variation of SpecAugment \cite{park2019specaugment} that uses the same masking
procedure as the contrastive loss is used for data augmentation in the ASR task.

\begin{table}[t]
\caption{Word error rates of models trained on the Librispeech 960-hours unlabeled and 100-hours labeled datasets.}
\label{low_resource}
\begin{small}
\begin{center}
\begin{tabular}{cccccc}
\toprule
\multirow{2}{*}{Method} & \multirow{2}{*}{LM} & \multicolumn{2}{c}{Dev} & \multicolumn{2}{c}{Test} \\
& & clean & other & clean & other \\
\midrule
Noisy student \cite{park2020improved} & LSTM & 3.9 & 8.8 & 4.2 & 8.6 \\
\midrule
wav2vec \textsc{Base}  & None & 6.1 & 13.5 & 6.1 & 13.3 \\
(quantized) \cite{baevski2020wav2vec} & 4-gram & 2.7 & 7.9 & 3.4 & 8.0 \\
 & Transf. & 2.6 & 7.0 & 2.9 & 6.8 \\
\midrule
wav2vec \textsc{Base}  & None & 6.0 & 14.3 & 6.1 & 14.6 \\
(continuous,  & 4-gram & 3.2 & 8.9 & 3.6 & 9.0 \\
reproduction) & Transf. & 1.9 & 8.1 & 3.1  & 7.9 \\
\midrule
Joint \textsc{Base} & None & 6.1 & 13.7 & 6.2 & 13.9 \\
(continuous) & 4-gram & 3.0 & 7.7 & 3.4 & 8.4 \\
 & Transf. & 2.1 & 6.4 & 2.7 & 6.8 \\
\bottomrule
\end{tabular}
\end{center}
\end{small}
\vspace{-0.7cm}
\end{table}

\begin{table}[t]
\caption{Word error rates of models trained on the Librispeech 960-hours unlabeled and 100-hours labeled datasets.}
\label{low_resource2}
\begin{small}
\begin{center}
\begin{tabular}{cccccc}
\toprule
\multirow{2}{*}{Method} & \multirow{2}{*}{LM} & \multicolumn{2}{c}{Dev} & \multicolumn{2}{c}{Test} \\
& & clean & other & clean & other \\
\midrule
Noisy student \cite{park2020improved} & LSTM & 3.9 & 8.8 & 4.2 & 8.6 \\
\midrule
wav2vec \textsc{Large}  & None & 4.6 & 9.3 & 4.7 & 9.0 \\
(quantized) \cite{baevski2020wav2vec} & 4-gram & 2.3 & 5.7 & 2.8 & 6.0 \\
 & Transf. & 2.1 & 4.8 & 2.3 & 5.0 \\
\midrule
Joint \textsc{Large} & None & 4.2 & 8.9 & 4.3 & 9.2 \\
(continuous) & 4-gram & 2.6 & 6.1 & 3.0 & 6.5 \\
 & Transf. & 2.0 & 5.1 & 2.5 & 5.3  \\
\bottomrule
\end{tabular}
\end{center}
\end{small}
\vspace{-0.7cm}
\end{table}

\subsection{Training}
The model is trained using the Adam optimizer (\cite{kingma2014adam}) for both losses with $\beta_1 = 0.9,\,\beta_2 = 0.98,\,\epsilon = 10^{-6}$ and weight decay $0.01$. The gradient for the convolutional encoder is scaled by $0.1$ for each of the two losses. The ratio of unsupervised to supervised loss updates is set to 1:1. The learning rate (LR) for the unsupervised loss is $5\times10^{-4}$ and for the supervised loss is $2.5\times10^{-5}$ for the \textsc{Base} model, whereas the LR for the unsupervised loss is $3\times10^{-4}$ and for the supervised loss is $2\times10^{-5}$ for \textsc{Large} model when using the 100-hours dataset as the labeled data. The LR for the unsupervised loss is $5\times10^{-4}$ and for the supervised loss is $1\times10^{-4}$ for the \textsc{Base} model when using the 960-hours dataset as the labeled data. 

The total number of updates is 500K. The LR for the both losses is warmed up from $0$ to their respective values in 20K updates. After the warmup period, the LR of the unsupervised loss $\eta_u$ is decayed to $0.1\eta_u$ at the end of training, whereas the LR of the supervised loss is kept constant. SpecAugment in
the supervised loss update is activated after the warmup period.

Training is performed on 64 V100 GPUs with a batch size per GPU equal to $87.5$s of audio for the \textsc{Base} model
and on 256 V100 GPUs with a batch size per GPU equal to $40$s of audio for the \textsc{Large} model. The audio samples are
batched together such that the total length of the samples does not exceed the batch size. 
The model is trained using the wav2letter++ toolkit~\cite{pratap2018wav2letter++} for approximately 4 days.

\subsection{Beam-search decoding and rescoring}
\label{sec:decoding}

Besides reporting word error rate (WER) without a language model (LM), we also perform a one-pass beam-search decoder with a 4-gram word-level LM~\cite{likhomanenko2019needs} and further the beam rescoring with a strong word-level Transformer LM~\cite{synnaeve2019end}. We rely on the beam-search decoder from the wav2letter++ toolkit~\cite{pratap2018wav2letter++} and follow the procedure from~\cite{synnaeve2019end}.

\section{Results and Discussion}

\subsection{Evaluation on standard SSL datasets}
The single-stage training pipeline is evaluated in a setting where there is a large
amount of unlabeled data compared to labeled data.

Table \ref{low_resource} shows word error rates (with and without an LM, see Section~\ref{sec:decoding}) for the \textsc{Base} model trained on Librispeech 960-hours unlabeled data and 100-hours labeled data. The joint training procedure generates an ASR model that 
matches the WER of the wav2vec 2.0 \textsc{Base} model on both the test-clean and test-other datasets. Unlike the wav2vec 2.0 model, this model
does not include quantization,
operates in the continuous space and does not use any unsupervised loss penalty terms
during training. 
Using the two-stage pipeline of wav2vec 2.0 (reproduced in wav2letter++) to train
the continuous \textsc{Base} model results in slightly worse ASR performance compared
to the quantized wav2vec 2.0 \textsc{Base} model.

Table \ref{low_resource2} shows word error rates (with and without an LM, see Section~\ref{sec:decoding}) for the \textsc{Large} model trained on Librispeech 960-hours unlabeled data and 100-hours labeled data. The joint training procedure generates an ASR model that 
matches the WER of the wav2vec 2.0 \textsc{Large} model on both the test-clean and test-other datasets. 

\subsection{Effect of 
hyperparameters on downstream task}
Table \ref{hyperparams} shows the effect of different
hyperparameters on the ASR performance of the model trained using the
single-stage training method. All models are trained for 500K updates
using the Librispeech 960-hours dataset as the unsupervised dataset and
the 100-hours dataset as the supervised dataset.
The baseline model uses a $\mathcal{L}_u$ to $\mathcal{L}_s$ update ratio equal to 1:1, $\mathcal{L}_u$ to 
$\mathcal{L}_s$ learning rate ratio equal to 20:1
and separate optimizers for each of the two losses.
Using a lower $\mathcal{L}_u$ to $\mathcal{L}_s$ 
learning rate ratio or using a single optimizer results in a higher WER on the dev-other dataset
compared to the baseline. The training pipeline is not sensitive to the update
ratio as can be seen by the negligible difference in WER between the models with
a $\mathcal{L}_u$ to $\mathcal{L}_s$ loss update ratio 1:1 and 5:1.

\begin{table}[h]
\caption{Word error rate (dev-other dataset, 4-gram LM) of models with different hyperparameters compared to baseline. }
\label{hyperparams}
\begin{small}
\begin{center}
\begin{tabular}{cccc}
\toprule
Hyperparameter & Updates & LR & dev-other \\
\midrule
Baseline &  1:1 & 20:1 & 8.0 \\
$\mathcal{L}_u$ to $\mathcal{L}_s$ update ratio & 5:1 & 20:1 & 7.9 \\
$\mathcal{L}_u$ to $\mathcal{L}_s$ learning rate ratio & 1:1 & 4:1 & 9.0 \\
Single optimizer & 1:1 & 20:1 & 11.1 \\ %
\bottomrule
\end{tabular}
\end{center}
\end{small}
\vspace{-0.7cm}
\end{table}

\subsection{Regularization effect %
on supervised loss
}
\begin{figure}[t]
\vspace{-0.5cm}
    \begin{center}
    \includegraphics[scale=0.44,height=5cm]{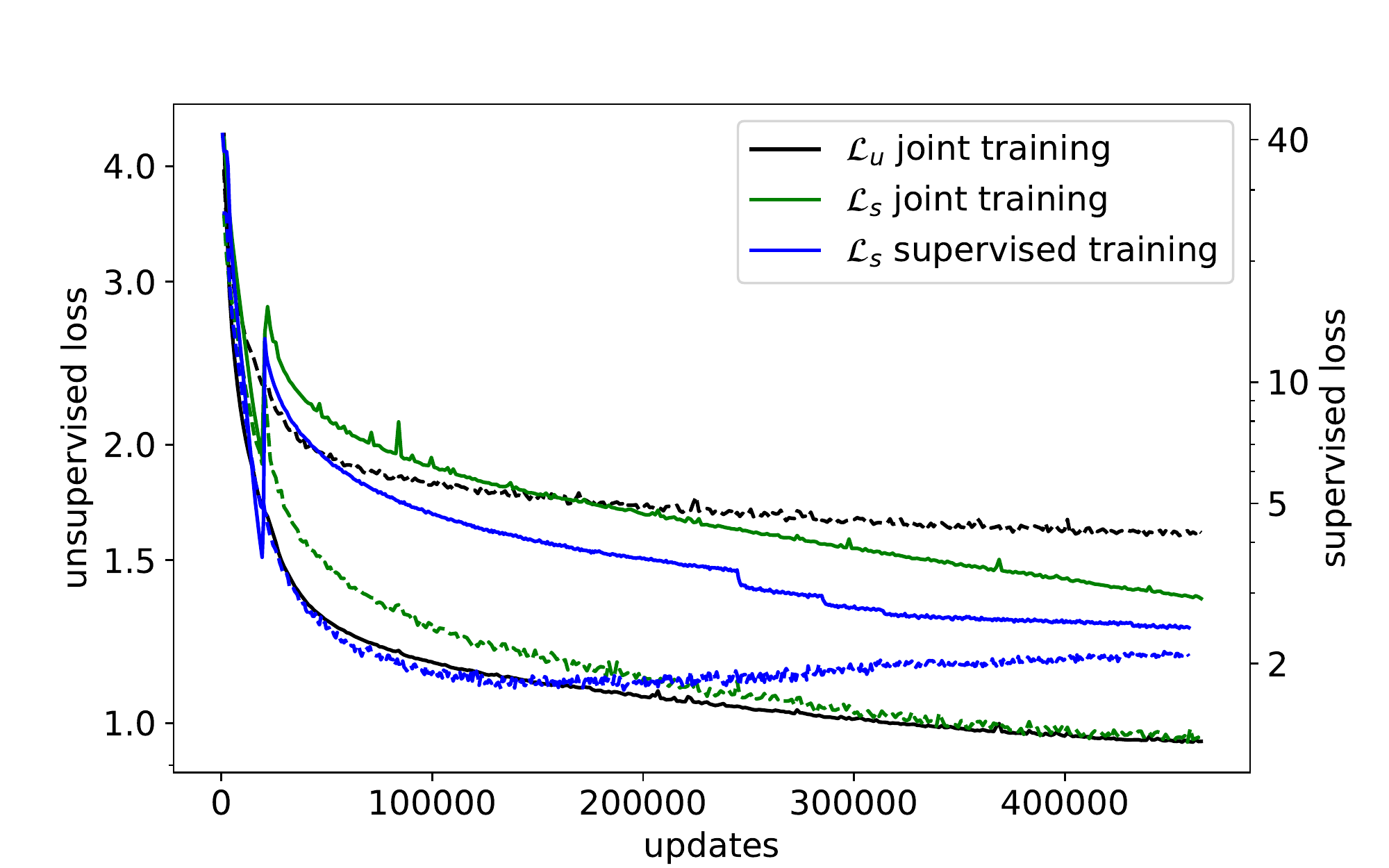}
    \caption{Unsupervised $\mathcal{L}_u$ and supervised $\mathcal{L}_s$ loss behaviour on the train (solid) and validation (dotted) sets for joint training ($\mathcal{L}_u$-black, $\mathcal{L}_s$-green) and supervised only training ($\mathcal{L}_s$-blue).
    }
    \label{f:correlation}
    \end{center}
\vspace{-0.5cm}
\end{figure}

Figure \ref{f:correlation} shows a plot of
the unsupervised loss $\mathcal{L}_u$ and
the supervised loss $\mathcal{L}_s$ on the train (Librispeech 960-hours) and validation (Librispeech dev-other) datasets
as a function of
total number of updates for the \textsc{Base} model trained using either  
joint training or supervised only training. 
Both models are trained for the same total number of updates, 500K. The supervised loss attains
a lower value on the validation dataset and a higher
value on the train dataset with joint training in comparison
to supervised only training. Furthermore, Table \ref{reg_effect} shows that 
a model trained using joint training achieves lower
WER (with and without an LM)
compared to a model trained using
supervised loss only, even though it has a lower number of updates from this loss.
This suggests that 
our method provides a regularizing
effect to the supervised loss.

\section{Related Work}
This paper draws upon recent advances in self-supervised contrastive learning \cite{oord2018representation,chen2020simple,misra2020self}. It uses
the principle of contrastive learning: similarity between an anchor and
positive samples is compared against similarity with negative samples.
But, the goal of
self-supervised learning is to learn representations that are useful for multiple 
downstream tasks. Whereas, our method is designed to maximize 
performance on a single downstream task.

More broadly, our single-stage training method can be linked to semi-supervised learning or self-training methods \cite{thomas2013deep,synnaeve2020endtoend,hsu2020semi,park2020improved,xu2020iterative} for ASR. 
These methods bootstrap an acoustic model (AM) from transcriptions (labeled data), transcribe  unlabeled audio with 
the trained AM (optionally with the help of an LM) and then
retrain the AM on the generated pseudo-labels. 
Self-training methods are complementary to our method
and there is potential to combine the two methods. 

As our approach addresses both, a contrastive learning task and speech recognition task, this paper 
is related to the field of multi-task learning \cite{kim2017joint,shinohara2016adversarial}. Recent
approaches to multi-task learning \cite{chen2015speech,he2018improved} solve the tasks by minimizing a loss, containing multiple terms, on the same supervised datasets. Whereas, in our method, 
the unsupervised and supervised losses are minimized
on their respective datasets.

\begin{table}[t]
\caption{Word error rates of models trained on Librispeech 960-hours labeled dataset.}
\label{reg_effect}
\begin{small}
\begin{center}
\begin{tabular}{cccccc}
\toprule 
\multirow{2}{*}{Method} & \multirow{2}{*}{LM} & \multicolumn{2}{c}{Dev} & \multicolumn{2}{c}{Test} \\
& & clean & other & clean & other \\
\midrule 
\multirow{3}{*}{Supervised} & None & 3.2 & 10.8 & 3.4 & 10.4 \\
 & 4-gram & 2.1 & 7.2 & 2.7 & 7.2 \\
 & Transf. & 1.5 & 5.4 & 2.2 & 5.6 \\
\midrule 
\multirow{3}{*}{Joint training} & None & 3.4 & 9.0 & 3.6 & 9.2 \\
 & 4-gram & 2.1 & 5.8 & 2.6 & 6.3 \\
 & Transf. & 1.5 & 4.4 & 2.1 & 4.8 \\
\bottomrule
\end{tabular}
\end{center}
\end{small}
\vspace{-0.7cm}
\end{table}

\section{Conclusion}
Our single-stage training method 
simplifies the process for learning speech recognition models jointly from
labeled and unlabeled data and allows directly optimizing the model on the downstream task.
Furthermore, the trained models match the performance of state of the
art self-supervised models for speech that use a two-stage pipeline. 
Finally, we demonstrate that solving the contrastive task provides
a regularizing effect on the supervised loss when only using a labeled dataset.

Finally, we would like to thank Alexei Baevski and Michael Auli for helpful discussions regarding wav2vec 2.0.
\newpage

\begin{small}
\bibliography{references}
\bibliographystyle{IEEE}
\end{small}

\end{document}